\pdfoutput=1
\documentclass[10pt,twocolumn,letterpaper]{article}

\usepackage[pagenumbers]{cvpr} 

\usepackage{graphicx}
\usepackage{amsmath}
\usepackage{amssymb}
\usepackage{booktabs}
\usepackage{rotating}
\usepackage{array}
\usepackage{colortbl}
\usepackage{multirow}
\usepackage{color}
\usepackage{enumitem}
\usepackage[dvipsnames]{xcolor}

\usepackage{bbm}
\usepackage{wrapfig}

\usepackage[pagebackref,breaklinks,colorlinks]{hyperref}

\usepackage[capitalize]{cleveref}
\crefname{section}{Sec.}{Secs.}
\Crefname{section}{Section}{Sections}
\Crefname{table}{Table}{Tables}
\crefname{table}{Tab.}{Tabs.}

\def\cvprPaperID{1399} 
\def\confName{CVPR}
\def\confYear{2023}

\begin{document}

\title{DynaMask: Dynamic Mask Selection for Instance Segmentation}

\author{Ruihuang Li\thanks{denotes the equal contribution, \dag denotes the corresponding author. This work is supported by the Hong Kong RGC RIF grant (R5001-18).},\quad Chenhang He\footnotemark[1],\quad Shuai Li,\quad Yabin Zhang,\quad Lei Zhang\footnotemark[2]\\
The Hong Kong Polytechnic
University\\
{\tt\small \{csrhli, csche, cslzhang\}@comp.polyu.edu.hk} }
\maketitle

\begin{abstract}
	The representative instance segmentation methods mostly segment different object instances with a mask of the fixed resolution, e.g., $28\times 28$ grid. However, a low-resolution mask loses rich details, while a high-resolution mask incurs quadratic computation overhead. It is a challenging task to predict the optimal binary mask for each instance. In this paper, we propose to dynamically select suitable masks for different object proposals. First, a dual-level Feature Pyramid Network (FPN) with adaptive feature aggregation is developed to gradually increase the mask grid resolution, ensuring high-quality segmentation of objects. Specifically, an efficient region-level top-down path (r-FPN) is introduced to incorporate complementary contextual and detailed information from different stages of image-level FPN (i-FPN). Then, to alleviate the increase of computation and memory costs caused by using large masks, we develop a Mask Switch Module (MSM) with negligible computational cost to select the most suitable mask resolution for each instance, achieving high efficiency while maintaining high segmentation accuracy. Without bells and whistles, the proposed method, namely DynaMask, brings consistent and noticeable performance improvements over other state-of-the-arts at a moderate computation overhead. The source code: \url{https://github.com/lslrh/DynaMask}.
\end{abstract}


\section{Introduction}
Instance segmentation (IS) is an important computer vision task, aiming at simultaneously predicting the class label and the binary mask for each instance of interest in an image. It works as the cornerstone of many downstream vision applications, such as autonomous driving, video surveillance, and robotics, to name a few. Recent years have witnessed the significant advances of deep convolutional neural networks (CNNs) based IS methods with a rich amount of training data as the rocket fuel \cite{bolya2019yolact,he2017mask,kirillov2020pointrend,tian2020conditional,shen2021dct}. Existing IS methods can be roughly divided into two major categories: two-stage~\cite{he2017mask,kirillov2020pointrend,cheng2020boundary} and single-stage methods~\cite{bolya2019yolact,chen2020blendmask,tian2020conditional}. The former first detect a sparse set of proposals and then performs mask predictions based on them, while the latter directly predict classification scores and masks based on the pre-defined anchors. Generally speaking, two-stage methods could produce higher segmentation accuracy but cost more computational resources than single-stage methods.  
\begin{figure}
	\centering 
	\includegraphics[scale=0.34]{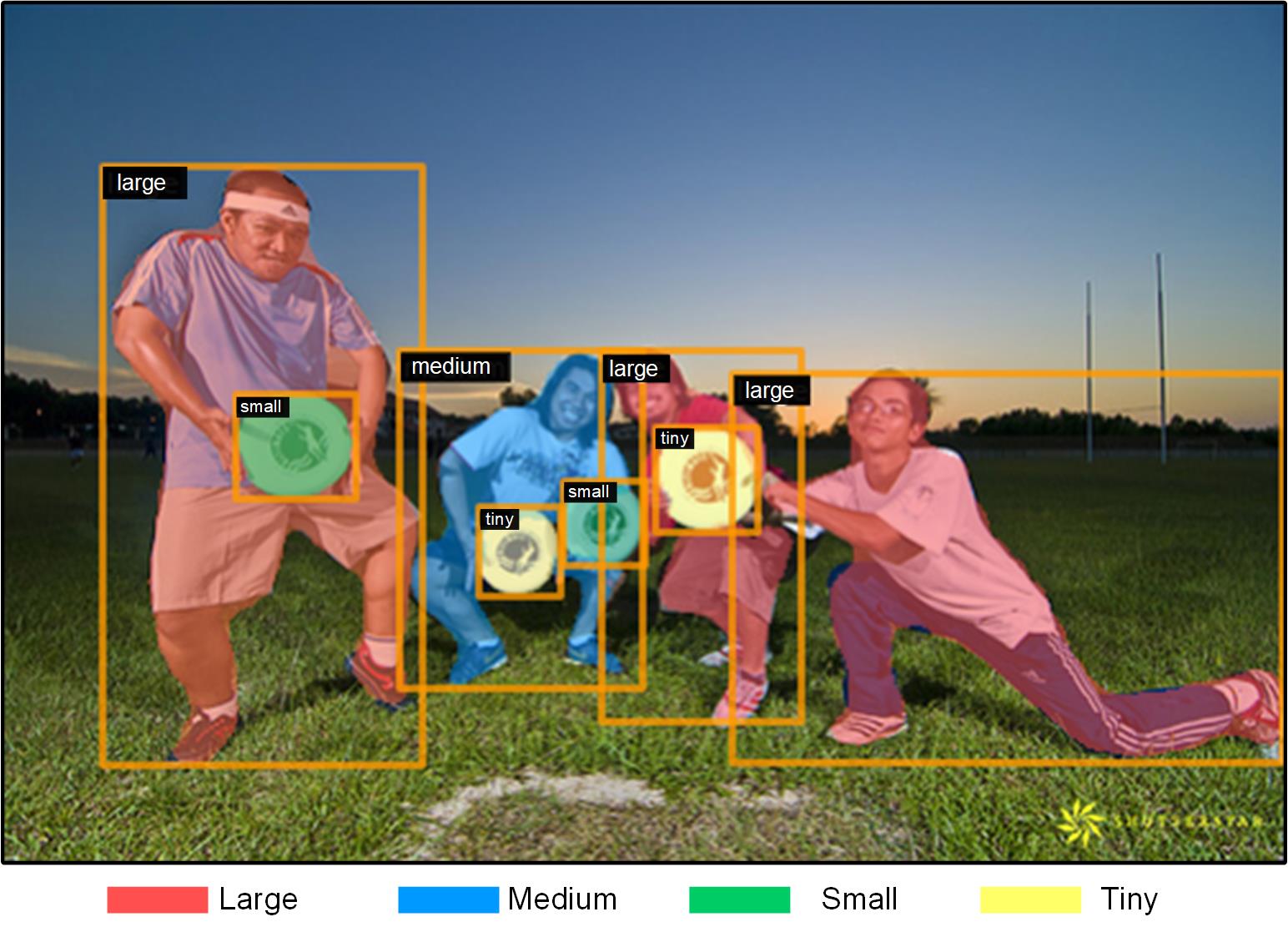}\\
	\vspace{-0.5em}
	\caption{Dynamic mask selection results. Some ``hard'' samples with irregular shapes like ``person'' are assigned larger masks, while the ``easy'' ones like ``frisbee'' are assigned smaller ones. }
	\vspace{-1.5em}
	\label{fig0}
\end{figure}

Among the many recently developed IS methods, the proposal-based two-stage methods~\cite{he2017mask,kirillov2020pointrend,cheng2020boundary}, which follow a detection-and-segmentation paradigm, are still the top performers. These methods need to predict a binary grid mask of uniform resolution for all proposals, \eg, $28\times 28$, and then upsample it to the original image size. For instance, Mask R-CNN \cite{he2017mask} first generates a group of proposals with an object detector and then performs per pixel foreground/background segmentation on the Regions of Interest (RoIs)~\cite{ren2015faster}. Despite achieving promising performance, the low-resolution mask of Mask R-CNN is insufficient to capture more detailed information, resulting in unsatisfactory predictions, especially over object boundaries. An intuitive solution to improve the segmentation quality is to adopt a larger mask. Nevertheless, high-resolution masks often generate excessive predictions on the smooth regions, resulting in high computational complexity. 

It is difficult to segment different objects in an image with masks of the same resolution. Objects with irregular shapes and complicated boundaries demand more fine-grained masks to predict, referred to as ``hard" samples, such as the ``person" in Fig.~\ref{fig0}. In comparison, the ``easy" samples with regular shapes and fewer details can be efficiently segmented using coarser masks, like the ``frisbee" in Fig.~\ref{fig0}. Inspired by the above observations, we propose to adaptively adjust the mask size for each instance for better IS performance. Specifically, instead of using a uniform resolution for all instances, we assign high-resolution masks to ``hard" objects and low-resolution masks to ``easy" objects. In this way, the redundant computation for ``easy" samples can be reduced while the accuracy of ``hard'' samples can be improved, achieving a balance between accuracy and speed. As shown in Tab.~\ref{tab0}, however, directly predicting a high-resolution mask by Mask R-CNN~\cite{he2017mask} unexpectedly degrades the mask average precision (AP). This attributes to two main reasons. First, the RoI features of larger objects are extracted from higher pyramid levels, which are very coarse due to the downsampling operations~\cite{lin2017feature}. Thus simply increasing the mask size of these RoIs will not bring extra useful information. Second, the mask head of Mask R-CNN is oversimplified, so it cannot make more precise predictions as the mask grid size increases. 

To overcome the above mentioned problems, we propose a dual-level FPN framework to gradually enlarge the mask grid. Specifically, in addition to traditional image-level FPN (i-FPN), a region-level FPN (r-FPN) is designed to achieve coarse-to-fine mask prediction. Wherein we construct information flows between i-FPN and r-FPN at different pyramid levels, aiming to incorporate complementary contextual and detailed information from multiple feature levels for high-quality segmentation. With the dual-level FPN, we present a data-dependent Mask Switch Module (MSM) with negligible computational cost to adaptively select masks for each instance. The overall approach, namely DynaMask, is evaluated on benchmark instance segmentation datasets to demonstrate its superiority over state-of-the-arts. 
The major contributions of this work are summarized as follows:
\begin{itemize}
	\item[$\bullet$] A dynamic mask selection method (DynaMask) is proposed to adaptively assign appropriate masks to different instances. Specifically, it assigns low-resolution masks to ``easy" samples for efficiency while assigning high-resolution masks to ``hard" samples for accuracy.\vspace{-0.5em}
	\item[$\bullet$] A dual-level FPN framework is developed for IS. We construct direct information flows from i-FPN to r-FPN at multiple levels, facilitating complementary information aggregation from multiple pyramid levels. \vspace{-0.5em}
	\item[$\bullet$] Extensive experiments demonstrate that DynaMask achieves a good trade-off between IS accuracy and efficiency, outperforming many state-of-the-art two-stage IS methods at a moderate computation overhead.  
\end{itemize}

\begin{table}[!t]
	\centering 
	\scalebox{0.9}{
		\begin{tabular}{c|ccc}
			\toprule \rowcolor{gray!20}
			Method                        & Resolution & AP   & FLOPs \\ \hline \hline
			\multirow{4}{*}{Mask R-CNN~\cite{he2017mask}}   &
			14$\times$14      & 32.9 & 0.2G  \\& 28$\times$28      & 34.7 & 0.5G  \\
			& 56$\times$56      & 33.8 & 2.0G  \\
			& 112$\times$112    & 32.5 & 8.0G  \\ \hline
			\multirow{4}{*}{DynaMask} 
			& 14$\times$14      & 32.9 & 0.2G\\
			& 28$\times$28      & 36.1 & 0.6G                          \\
			& 56$\times$56      & 37.1 & 1.0G  \\
			& 112$\times$112    & 37.6 & 1.4G  \\ \bottomrule
	\end{tabular}}
	\caption{Mask AP and FLOPs with different mask resolutions. For Mask R-CNN, directly increasing the mask resolution will decrease the mask AP. While for our DynaMask, higher mask resolution results in better performance. }
	\vspace{-1em}
	\label{tab0}
\end{table}

\section{Related Work}
\textbf{Instance Segmentation.} To date, most of the top-performing IS methods still follow the Mask R-CNN meta-architecture~\cite{he2017mask}. These proposal-based approaches typically employ an object detector to localize each instance in bounding boxes. Then the instance-wise features are cropped and extracted from FPN features based on the detected bounding boxes by using RoI pooling/align~\cite{ren2015faster,he2017mask}. Finally, a compact segmentation head is deployed to obtain the desired object masks. Mask Scoring R-CNN ~\cite{huang2019mask} aligns the mask quality and score by using a branch to explicitly learn the quality of predicted masks. BMask R-CNN~\cite{cheng2020boundary} leverages boundary details to improve mask localization ability. DCT-Mask~\cite{shen2021dct} encodes high-resolution binary masks into compact vectors through the discrete cosine transform (DCT). PANet~\cite{liu2018path} constructs two feature pyramids for improving mask prediction. 

\begin{figure*}[!t]
	\centering
	\includegraphics[scale=0.85]{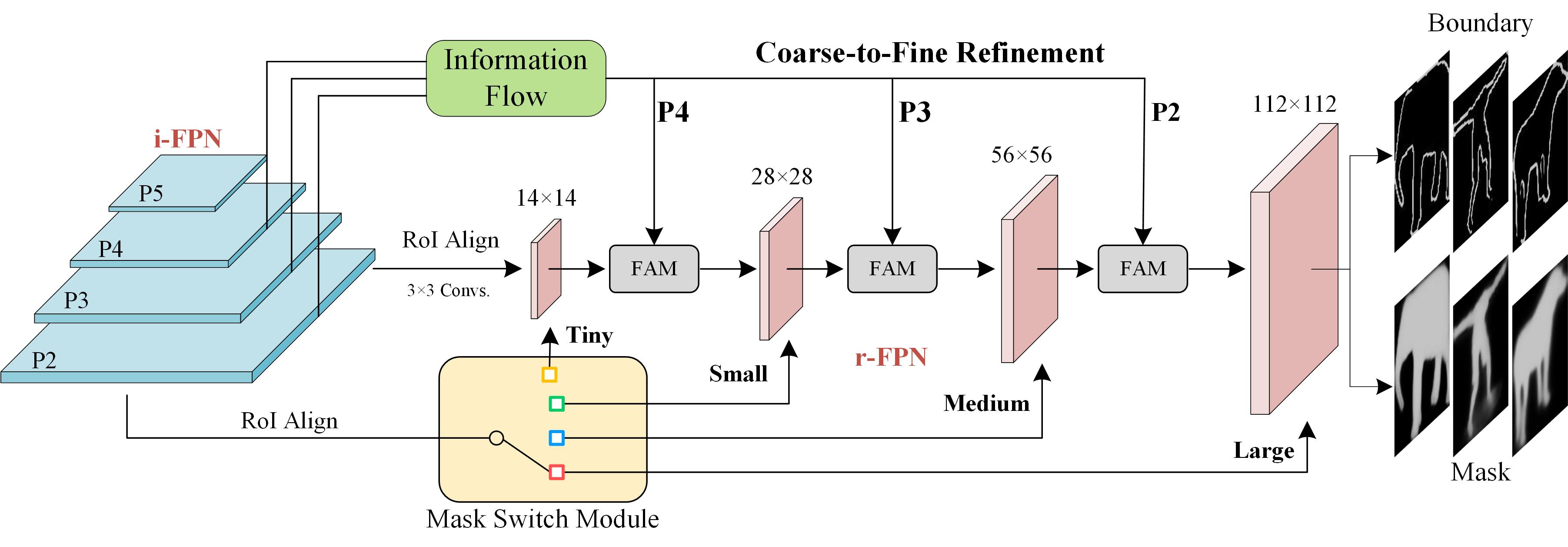}\\ \vspace{-0.5em}
	\caption{The overall architecture of DynaMask. We propose a dual-level FPN framework to gradually increase the mask size and achieve higher-quality IS. The information flow is constructed between each level of image-level FPN (i-FPN) and region-level FPN (r-FPN), so that region-wise feature hierarchies $\{L_{tiny}, L_{small}, L_{medium}, L_{large}\}$ are constantly enhanced by incorporating complementary information from $\{P2,P3,P4\}$ of i-FPN, resulting in coarse-to-fine mask predictions. To reduce the computation and memory cost, a Mask Switch Module (MSM) is developed to predict the mask resolution for each instance with budgeted resource consumption. Specifically, MSM outputs four different switching states, corresponding to four different mask resolutions, \textit{i.e.}, $[14\times 14, 28\times 28, 56\times 56, 112\times 112]$.}
	\label{fig1}
	\vspace{-1em}
\end{figure*}
Some works~\cite{chen2019hybrid,kirillov2020pointrend,zhang2021refinemask} have been proposed to improve the mask quality by coarse-to-fine refinement. HTC~\cite{chen2019hybrid} performs cascaded refinement on both detection and segmentation tasks and explores the reciprocal relationships between them. PointRend~\cite{kirillov2020pointrend} and RefineMask~\cite{zhang2021refinemask} extract fine-grained features in a multi-stage manner. The former performs point-based predictions over the blurred areas, while the latter refines the entire RoI feature. Despite the promising segmentation results, multiple refinement stages inevitably increase the inference time and memory burden. In this paper, we dynamically select the suitable mask for each instance so that ``easy'' samples are assigned small masks (with fewer refinement stages) for efficiency, and ``hard'' samples are assigned large masks (with more refinement stages) for accuracy, achieving a balance between efficiency and accuracy.

\textbf{Dynamic Networks.} Dynamic networks, which aim to adaptively adjust the network architecture according to the input feature, have been widely studied in recent years. Dynamic model compression methods either drop blocks~\cite{huang2018multi,mullapudi2018hydranets,wang2018skipnet} or prune channels~\cite{lin2017runtime,you2019gate} for speeding up the inference. For instance, SkipNet~\cite{wang2018skipnet} skips convolutional blocks through a reinforcement learning-based gating network. Huang \etal propose a multi-scale dense network with multiple classifiers for allocating different computations for ``easy" and ``hard" samples. Li \etal~\cite{li2020learning} adopt an end-to-end dynamic routing framework to alleviate the scale variance among inputs. DRNet~\cite{zhu2021dynamic} attempts to reduce the redundancy on the input resolution of modern CNNs. 

However, employing dynamic masks at different resolutions for segmenting different instances has rarely been explored in the field of IS. Conventional methods~\cite{cheng2020boundary,he2017mask,kirillov2020pointrend,li2022class,li2021t} typically predict a fixed-size mask irrespective of the object type. This is sufficient for ``easy" samples but produces over-smoothing predictions for ``hard" samples over the fine-level details. In order to improve the segmentation performance without introducing many additional computation burdens, we devise a dynamic mask selection framework to adaptively allocate suitable masks to different objects according to their segmentation difficulties.    

\section{Dynamic Mask Selection}
The framework of DynaMask is illustrated in Fig.~\ref{fig1}. A dual-level FPN architecture is first proposed for improving IS quality, and then a Mask Switch Module (MSM) is developed to dynamically allocate appropriate masks to each instance, so that the resource consumption can be reduced while maintaining superior IS performance. Our DynaMask produces high-quality segmentation with a moderate computation overhead. 

\subsection{Dual-Level FPN}
Original image-level FPN (i-FPN)~\cite{lin2017feature} introduces a top-down path to propagate contextual semantic information from higher layers to lower ones. Actually, the lower-level features contain more fine-grained details than higher-level ones, which are beneficial for high-quality segmentation, especially on boundary regions, but these information is not fully explored in Mask R-CNN~\cite{he2017mask}. In this work, we propose a region-level FPN (r-FPN) to integrate more detailed information from lower layers of i-FPN into region-wise feature hierarchies. The information flows from each level of i-FPN to r-FPN are shown in Fig.~\ref{fig1}.   

\noindent\textbf{Region-Level FPN.} We follow the original i-FPN to define the layers producing feature maps of the same resolution as one network stage corresponding to one feature level. We use $\{P_2, P_3, P_4, P_5\}$ to denote different feature levels of i-FPN. The r-FPN starts from an RoI-Aligned region-wise feature, and it is gradually enhanced by fusing the complementary information from $\{P_2, P_3, P_4\}$ of i-FPN, resulting in top-down region-based feature hierarchies, denoted by $\{L_{tiny}, L_{small}, L_{medium}, L_{large}\}$. From $L_{tiny}$ to $L_{large}$, the spatial resolution is progressively increased by a factor of two. We design a Feature Aggregation Module (FAM) to integrate the r-FPN feature $L_r$ and the i-FPN feature $P_i$. 

\begin{figure}
	\centering
	\includegraphics[width=0.46\textwidth]{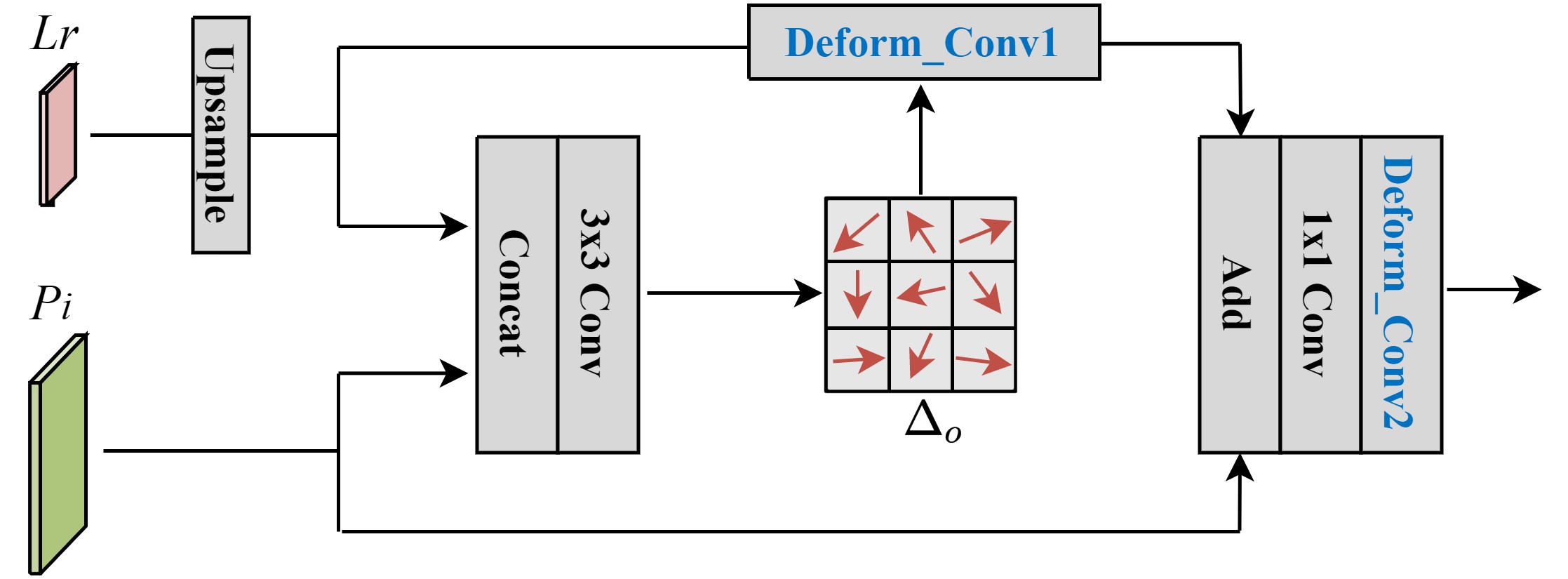}\\
	\caption{Feature Alignment Moudule (FAM). Deform\_Conv1 adjusts the spatial positions of upsampled r-FPN feature $L_{r}$ and the cropped i-FPN feature $P_i$. Deform\_Conv2 attends to the salient parts of objects.}
	\label{fig2}
	\vspace{-1.5em}
\end{figure}

\noindent\textbf{Feature Aggregation Module (FAM).} There exists spatial misalignments between $L_r$ and $P_i$ due to the upsampling and RoI pooling~\cite{ren2015faster} operations, which may degrade the segmentation performance on boundary areas. To overcome this limitation, we propose the FAM to adaptively aggregate multi-scale features. As show in Fig.~\ref{fig2}, FAM contains two deformable convolutions~\cite{dai2017deformable} which play different roles. The first one (\textit{Deform\_Conv1}) adjusts the positions of $L_r$, making it better aligned to $P_i$. Here we first concatenate $L_r$ with $P_i$, and then pass the concatenated features through a $3\times3$ \textit{Conv} to obtain the offset map, denoted by $\Delta_o$. Finally, $L_r$ is aligned to $P_i$ with the learned offset $\Delta_o$. The second one (\textit{Deform\_Conv2}) works like an attention mechanism, which attends to the salient parts of objects. The proposed FAM is plugged into different stages of r-FPN and it plays a key role in improving the mask prediction. 
  
\subsection{Mask Switch Module (MSM)}
The proposed dual-level FPN framework brings a significant performance improvement but at the price of expensive computation and memory burdens. Inspired by the fact that different instances require different mask grids to achieve accurate segmentation, we propose a novel method to adaptively adjust the mask grid resolution for different instances. Specifically, an MSM is developed to perform mask resolution prediction under a budgeted computation consumption, achieving a good trade-off between segmentation accuracy and efficiency. 

\begin{figure*}[!t]
	\vspace{1em}
	\centering
	\includegraphics[scale=0.8]{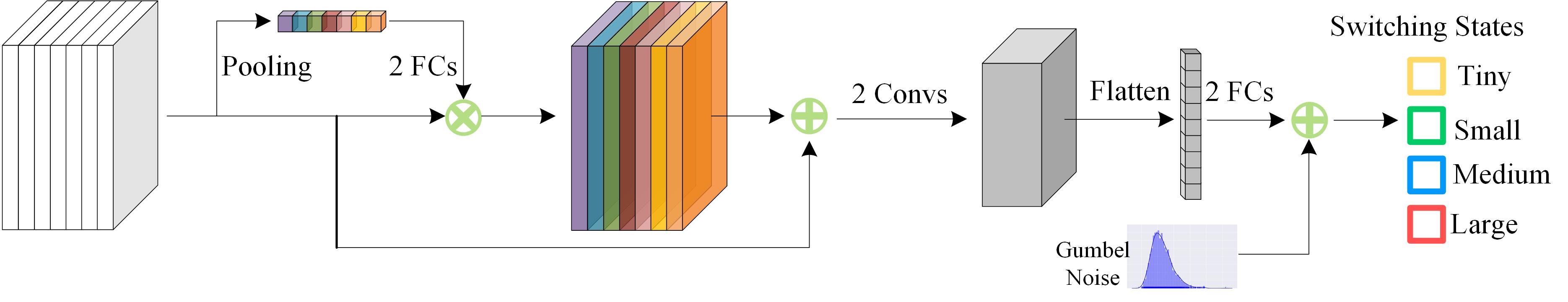}\\
	\vspace{-1em}
	\caption{The Mask Switch Module (MSM), which is a CNN-based classifier and consists of a SE-block~\cite{hu2018squeeze}, two convolutional layers and two fully-connected layers.}
	\label{fig3}
	\vspace{-0.5em}
\end{figure*}

\noindent\textbf{Optimal Mask Assignment.} The MSM module is actually a lightweight classifier, denoted by $f_{MSM}(\cdot)$, which is illustrated in Fig.~\ref{fig3}. It contains a channel-wise attention module~\cite{hu2018squeeze}, followed by a few convolutional layers and fully-connected layers. This classifier aims to find the optimal mask resolution from a collection of $K$ candidates,~\textit{i.e.}, $[r^1,r^2,\cdots,r^K]$, so that the instance could be accurately segmented with the minimal resource cost. Specifically, MSM takes the cropped region-wise RoI feature as input and outputs a probability vector $P=[p^1,\cdots,p^K]$ by taking a softmax operation. Each element of this vector represents the probability that the corresponding candidate resolution is selected:
\begin{align}
	p^k = \frac{\exp(f^k_{MSM}(x))}{\sum_{k'}\exp(f^{k'}_{MSM}(x))}, \quad k\in\{1,\cdots,K\},
\end{align}
where $x$ is the input RoI feature fed to MSM. The candidate resolution of the largest probability is chosen as the switching state, which decides the resolution of mask grid to segment an object. 

\noindent\textbf{Reparameterization with Gumbel-Softmax.} The soft output $P$ of MSM should be transformed into a one-hot prediction, denoted by $Y=[y^1,\cdots, y^K],\ y^k\in\{0,1\} $. This process can be done by discrete sampling, which however is non-differentiable and does not support end-to-end training. To allow the gradient back-propagation for updating MSM, we introduce a reparameterization method~\cite{jang2017categorical}, called the Gumbel-Softmax. Given a categorical distribution with class probabilities $P=[p^1,\cdots,p^K]$, we can draw a group of masks through the rule $y=\textrm{one\_hot}\Big({argmax_k}(logp^k+g^k)\Big)$, where $\{g^k\}_{i=k}^K$ are i.i.d. samples drawn from the $Gumbel(0,1)$ distribution, which is defined by $g=-log\big(-log(u)\big)\ \textrm{with}\ u\sim{Uniform}(0,1)$. 
Then we use the Gumbel-softmax function as a continuous, differentiable approximation to the original $\rm{softmax}$ function:
\begin{align}
	y^k=\frac{\exp\big((logp^k+g^k)/\tau\big)}{\sum_{k'} \exp\big((logp^{k'}+g^{k'})/\tau\big)},
\end{align}
where $\tau$ denotes a temperature parameter. When $\tau$ approches 0, the Gumbel-softmax is close to one-hot.

\subsection{Objective Function} 
The proposed framework gradually enlarges the mask resolution by a factor of two to improve the segmentation performance. On the one hand, FAM adaptively aggregates complementary information from multiple stages of i-FPN and r-FPN to enhance the feature hierarchies. On the other hand, MSM dynamically allocates masks of different resolutions for different instances in an image, reducing resource costs without sacrificing accuracy. In this subsection, we elaborate on the loss function used to train the mask head.    

\noindent\textbf{Mask Loss.} Given a positive instance $x_i$, we first predict its mask switching state $Y=[y^1,\cdots,y^K]$ by MSM, and obtain a group of mask prediction maps at $K$ different resolutions $\{\hat{m}_i^1, \cdots, \hat{m}_i^K\}$ by passing this instance through different stages of r-FPN. We define the mask loss function as follows:
\begin{align}
	\mathcal{L}_{mask}=\sum_{i=1}^N\sum_{k=1}^{K}y^k\ell(\hat{m}^k_i, m_i),
	\label{eq5}
\end{align}
where $\hat{m}^k_i$ denotes the $k$-th mask prediction of $x_i$, and $m_i$ represents its corresponding ground truth mask grid. $y^k$ is the indicator for whether the $k$-th mask resolution is selected as the output resolution. $\ell$ is defined as the binary cross-entropy loss in this paper. 

\begin{figure}[!t]
	\centering
	\includegraphics[width=0.38\textwidth]{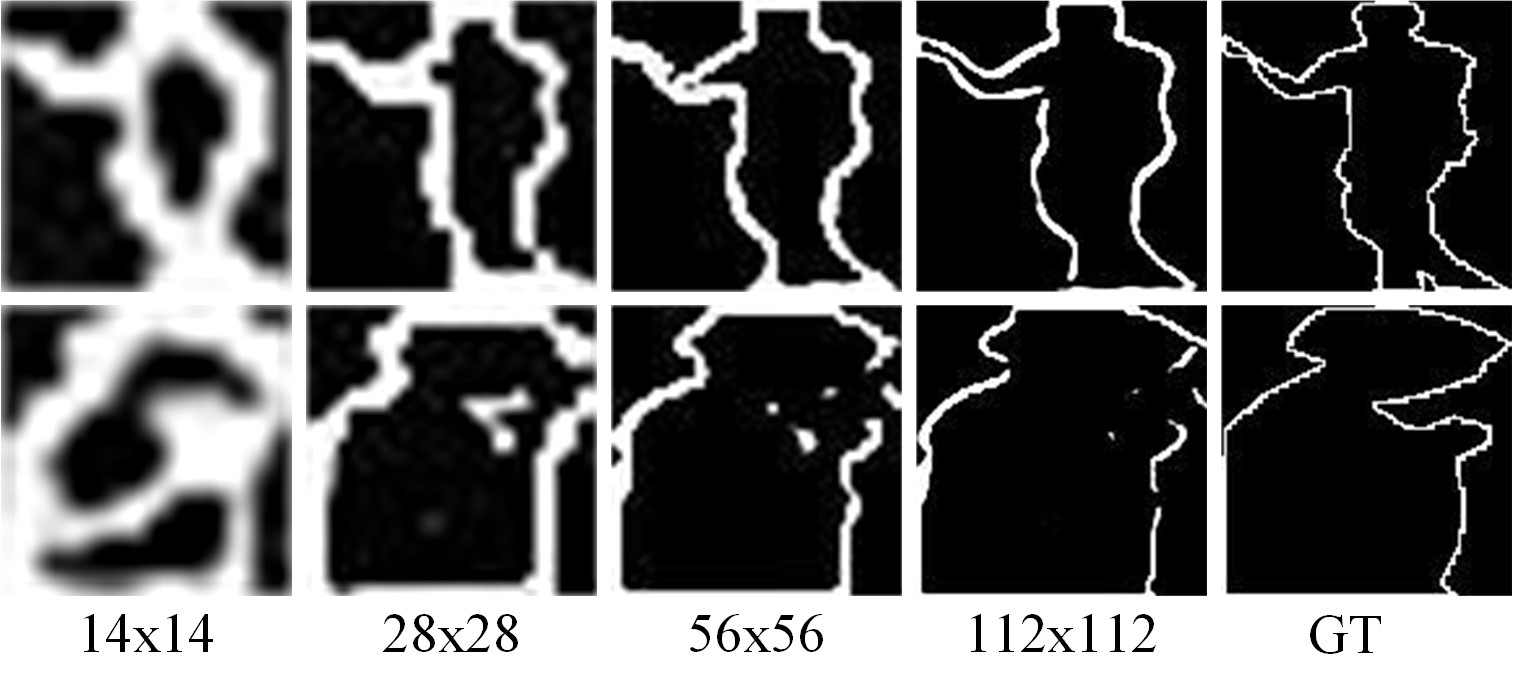}\\
	\vspace{-1em}
	\caption{The predicted object edges using the masks of different resolutions. The edge loss computed on boundary regions could reflect the quality of mask.}
	\vspace{-0.5em}
	\label{fig4}
\end{figure}

\noindent\textbf{Edge Loss.} In Eq.~\ref{eq5}, we assume that the mask producing smaller loss should have higher quality, so that the most accurate mask could be selected by minimizing the mask loss. However, our empirical results show that the mask loss produced on different masks is very close, making it difficult to distinguish the mask quality. In comparison, as shown in Fig.~\ref{fig4}, the edge loss produced by masks of different resolutions varies greatly, which could better reveal the mask quality.
Given the output of MSM $Y=[y^1,\cdots,y^K]$ and the edge maps of different resolutions, denoted by $\{\hat{e}^1_i, \cdots,\hat{e}^K_i \}$, the edge loss is defined as follows: 
\begin{align}
	\mathcal{L}_{edge}=\sum_{i=1}^N\sum_{k=1}^{K}y^k\ell(\hat{e}^k_i, e_i),
	\label{eq6}
\end{align}
where $e_i$ denotes the ground-truth edge, which is generated by first applying the Laplacian operator on the ground-truth mask $m_i$ to obtain a soft edge map, and then converting it into a binary edge map by thresholding~\cite{cheng2020boundary}. Fig.~\ref{fig4} visualizes the object edges predicted by using masks of different resolutions. As can be seen, the edge loss could better reveal the quality of masks, \ie, higher-resolution masks produce edges closer to GT (smaller loss), while lower-resolution masks produce edges more different from GT (larger loss).

\begin{figure*}[t]
	\centering
	\includegraphics[width=1.03\textwidth]{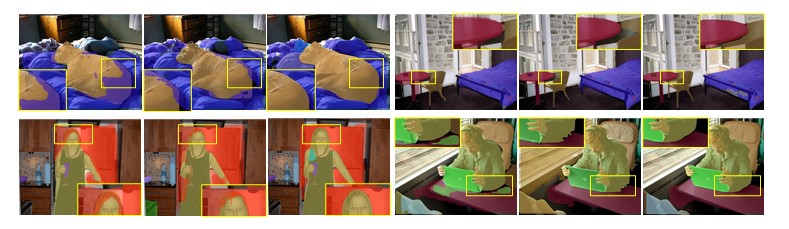}\\
	\vspace{-1.0em}
	\caption{Example results of Mask R-CNN~\cite{he2017mask} (left), PointRend~\cite{kirillov2020pointrend} (middle), and DynaMask (right), using ResNet-50-FPN as backbone.}
	\label{fig6}
	\vspace{-0.5em}
\end{figure*}

\noindent\textbf{Budget Constraint.} By optimizing the edge loss in Eq.~\ref{eq6}, the model tends to converge to a sub-optimal solution that all instances are segmented with the largest mask, \textit{i.e.}, $112\times 112$, which incorporates more detailed information and hence has the minimal prediction loss. Actually, not all samples require the largest mask for segmentation. The redundant computations of segmenting the ``easy" samples can be saved for efficiency. In order to reduce the computational cost and avoid the above mentioned sub-optimal solution, we propose to train the MSM with a budget constraint. Specifically, let $\mathcal{C}$ denote the computational cost (\textit{e.g.}, FLOPs) corresponding to the selected mask resolution. We add a penalty to the model when the expectation of FLOPs computed on current batch data, denoted by $\mathbb{E}(\mathcal{C})$, exceeds the target FLOPs, denoted by $\mathcal{C}_{t}$. The budget constraint is defined as follows:
\begin{align}
	\mathcal{L}_{budget}=\textrm{max}(\frac{\mathbb{E}(\mathcal{C})}{\mathcal{C}_{t}}-1, 0).
	\label{eq7}
\end{align}
We further introduce an information entropy loss to balance the resolution predictions of MSM. Given a group of output probability vectors $P_1,P_2,\cdots,P_N$, where $N$ is the number of instances of the current batch, the frequency of the $k$-th resolution is calculated by: $f^k=\frac{1}{N}\sum_i p_i^k$. Then the information entropy loss is defined as follows:
\begin{align}
	\mathcal{L}_{entropy}=\frac{1}{K}\sum_k f^k logf^k.
	\label{entropy}
\end{align}   
The above entropy loss tends to push each element $f^k$ to be $\frac{1}{K}$ so that MSM could select different resolutions with similar probabilities. 

Finally, the total objective function of the mask branch is obtained as follows:
\begin{align}
	\mathcal{L}_{total} = \mathcal{L}_{mask} + \lambda_1\mathcal{L}_{edge} + \lambda_2\mathcal{L}_{reg},
\end{align}
where $\lambda_1$ and $\lambda_2$ are the trade-off hyper-parameters. $\mathcal{L}_{reg}$ denotes the regularization term which is obtained by combining the budget constraint in Eq.~\ref{eq7} and the information entropy loss in Eq.~\ref{entropy}, \textit{i.e.}, $\mathcal{L}_{reg}=\mathcal{L}_{budget}+\mathcal{L}_{entropy}$.  

\section{Experiments}
We perform extensive experiments on the benchmark instance segmentation datasets: COCO~\cite{lin2014microsoft} and Cityscapes~\cite{cordts2016cityscapes}. COCO~\cite{lin2014microsoft} contains about 115k images (\texttt{train2017}) with instance-level annotations of 80 categories for training. We use the \texttt{val2017} set (around 5k images) for the ablation study and the \texttt{test-dev2017} set (about 20k images) for comparison with other methods. The Cityscapes dataset~\cite{cordts2016cityscapes} contains 2975, 500, and 1525 images collected from 8 categories of urban scenes for training, validation, and testing, respectively. On both COCO and Cityscapes datasets, we use the standard mask AP as the evaluation metric, which computes the average precision over varying IoU thresholds (from 0.5 to 0.95).

\subsection{Implementation Details}
We adopt Mask R-CNN~\cite{he2017mask} as our baseline. The backbone is pre-trained on ImageNet. The hyper-parameters and loss functions are set the same as Mask R-CNN implemented in MMDetection~\cite{chen2019mmdetection} unless specified. The proposed MSM has four switching states, which correspond to four candidate resolutions, \textit{i.e.}, $[14\times14, 28\times28, 56\times56, 112\times112]$. The hyper-parameter $\lambda_1$ is set as $0.1$. We first pre-train the model without MSM using mask loss at all resolutions for one epoch. The initial learning rate is 0.02, and the batch size is 16 on 8 GPUs. Then we train all modules for 12 epochs using SGD with the same initial learning rate and batch size and reduce the learning rate by a factor of 0.1 after 8 and 11 epochs, respectively. Multi-scale training is used with the shorter side randomly sampled from $[640, 800]$, while for inference, the short side is resized to 800. In the ablation study, we use the standard $1\times$ training schedule and data augmentation defined in MMDetection~\cite{chen2019mmdetection}.     

\noindent\subsection{Main Results}
\noindent\textbf{Comparison with Mask R-CNN.} We first compare the performance of DynaMask with the baseline Mask R-CNN on COCO by using ResNet-50 and ResNet-101 backbones. As shown in Tab.~\ref{tab1}, our method outperforms Mask R-CNN by a large margin. The performance is improved by 2.9\% AP and 2.8\% AP with ``$1\times$" and ``$2\times$" schedules, respectively, when ResNet-50-FPN backbone is used. Particularly, DynaMask outperforms the baseline by 3.3\%, and 3.6\% with ``$1\times$" and ``$2\times$" schedules in terms of $\rm AP_{75}$, respectively. Similar observations can be obtained when ResNet-101-FPN is used as the backbone. This is because the proposed dual-level FPN structure incorporates complementary semantic and detailed information from multiple levels of FPN, resulting in more precise mask localization and higher-quality segmentation.

\begin{table*}[!t]
	\centering
		\scalebox{0.85}{
			\begin{tabular}{c|c|c|c|cc|ccc}
				\toprule\rowcolor{gray!20}
				Method       & Backbone    & Sched.  & AP & $\rm AP_{50}$ & $\rm AP_{75}$ & $\rm AP_S$ & $\rm AP_M$ & $\rm AP_L$ \\ \hline\hline
				Mask R-CNN~\cite{he2017mask}   & {ResNet-50-FPN}   & 1$\times$ & 34.7   & 55.7  & 37.2    &  18.3   &  37.4  & 47.2       \\
				DynaMask & {ResNet-50-FPN}   & 1$\times$ & 37.6  & 57.4  &  40.5    &  20.7   &  40.4     &  50.3  \\ \hline
				Mask R-CNN~\cite{he2017mask}   & {ResNet-50-FPN}   & 2$\times$ &  35.4  & 56.4  & 37.9  &  19.1  &  38.6  &  48.4    \\
				DynaMask & {ResNet-50-FPN}   & 2$\times$ & 38.2  & 58.1  &  41.5    & 20.5  &    40.8   &  52.7    \\ \hline
				Mask R-CNN~\cite{he2017mask}   & {ResNet-101-FPN}  & 1$\times$ & 36.1  & 57.5  &  38.6    & 18.8  &    39.7   &  49.5       \\
				DynaMask & {ResNet-101-FPN}  & 1$\times$ & 38.7  & 58.8   & 41.8    & 20.9  &    41.8   &  52.4   \\ \hline
				Mask R-CNN~\cite{he2017mask}   & {ResNet-101-FPN}  & 2$\times$ & 36.6  &  57.9  & 39.1   & 19.2      &  40.2     &   50.5    \\
				DynaMask & {ResNet-101-FPN}  & $2\times$ & 39.0  &  59.1 
				& 42.2    &   20.9    &    42.1   & 53.3      \\ \bottomrule
		\end{tabular}}
		\vspace{-0.7em}
		\caption{Comparison with Mask R-CNN on COCO \texttt{val2017}.}
		\vspace{-0.5em}
		\label{tab1}
	\end{table*}
	
	\begin{table*}[!t]
		\centering
		\scalebox{0.8}{
			\begin{tabular}{c|c|c|c|cc|ccc}
				\toprule\rowcolor{gray!20}
				\multicolumn{1}{l|}{Method}             & \multicolumn{1}{c|}{Backbone}       & \multicolumn{1}{c|}{sched.} & AP   & $\rm AP_{50}$ & $\rm AP_{75}$ & $\rm AP_S$ & $\rm AP_M$ & $\rm AP_L$ \\ \hline\hline
				\multicolumn{1}{l|}{MEInst~\cite{zhang2020mask}}             & {ResNet-101-FPN} & 3$\times$     & 33.9 & 56.2      & 35.4      & 19.8   & 36.1   & 42.3   \\
				\multicolumn{1}{l|}{Mask R-CNN~\cite{he2017mask}}   & {ResNet-101-FPN}  & 3$\times$ &     38.5 & 60.0 & 41.6 & 19.2 & 41.6 & 55.8    \\
				\multicolumn{1}{l|}{BMask R-CNN~\cite{cheng2020boundary}}        & {ResNet-101-FPN} & 1$\times$     & 37.7 & 59.3      & 40.6      & 16.8   & 39.9   & 54.6   \\
				\multicolumn{1}{l|}{TensorMask~\cite{chen2019tensormask}}         & {ResNet-101-FPN} & 6$\times$      & 37.1 & 59.3      & 39.4      & 17.4   & 39.1   & 51.6   \\
				\multicolumn{1}{l|}{Mask Scoring~\cite{huang2019mask}} & {ResNet-101-FPN} & 18e      & 38.3 & 58.8      & 41.5      & 17.8   & 40.4   & 54.4   \\
				\multicolumn{1}{l|}{CondInst~\cite{tian2020conditional}}           & {ResNet-101-FPN}           & 3$\times$     & 39.1 & 60.9      & 42.0      & 21.5   & 41.7   & 50.9   \\
				\multicolumn{1}{l|}{BlendMask~\cite{chen2020blendmask}}          & {ResNet-101-FPN}               & 3$\times$     & 38.4 & 60.7      & 41.3      & 18.2   & 41.5   & 53.3   \\
				\multicolumn{1}{l|}{PointRend~\cite{kirillov2020pointrend}}          &  {ResNeXt-101-FPN}            & 3$\times$      & 41.4 & 63.3      & 44.8      & 24.2   & 43.9   & 53.2   \\
				\multicolumn{1}{l|}{SOLOv2~\cite{wang2020solov2}}             & {ResNet-101-FPN}               & 6$\times$     & 39.7 & 60.7      & 42.9      & 17.3   & 42.9   & 57.4   \\
				\multicolumn{1}{l|}{HTC~\cite{chen2019hybrid}}                & {ResNet-101-FPN}               & 20e      & 39.7 & 61.8      & 43.1      & 21.0   & 42.2   & 53.5   \\ 
				\multicolumn{1}{l|}{DCT-Mask$^\dag$~\cite{shen2021dct}}           & {ResNet-101-FPN}               & 3$\times$      & 40.1 &  61.2   &  43.6  & 22.7  & 42.7   & 51.8    \\ 
				\multicolumn{1}{l|}{RefineMask~\cite{zhang2021refinemask}}           & {ResNet-101-FPN}               & 3$\times$      & 39.4 &  -   &  -  &  - & -   & -  \\ 
				\multicolumn{1}{l|}{Mask Transfiner$^\dag$~\cite{ke2022mask}}           & {ResNet-101-FPN}               & 3$\times$      & 40.7 &  -   &  -  &  23.1  & 42.8   & 53.8    \\ 
				\multicolumn{1}{l|}{QueryInst$^\dag$~\cite{fang2021instances}}           & {ResNet-101-FPN}               & 3$\times$      & 41.7 &  64.4   &  45.3  &  24.2  & 43.9   & 53.9    \\ 
				\hline
				\multicolumn{1}{l|}{DynaMask}               &    {ResNet-101-FPN}                          &  3$\times$      & 39.6     &    61.4       &      42.9     &  21.2      &     42.4   &  53.2       \\ 
				\multicolumn{1}{l|}{DynaMask$^\dag$}               &    {ResNet-101-FPN}                          &  3$\times$      & 41.3     &    62.5       &      45.2     &  22.8      &     43.4   &  54.0       \\ 
				\multicolumn{1}{l|}{DynaMask$^\dag$}               &  {ResNeXt-101-FPN }                        &  3$\times$      & 42.0     &   62.9        &      46.3     &  23.4      &     44.0   &    54.8   \\ 
				\bottomrule
		\end{tabular}}
		\vspace{-0.7em}
		\caption{Comparison with state-of-the-art methods for instance segmentation on COCO \texttt{test-dev2017}, $\dag$ denotes multi-scale training.}
		\vspace{-1.0em}
		\label{tab6}
	\end{table*}
	
	\begin{figure*}[t]
		\centering
		\includegraphics[width=1.0\textwidth]{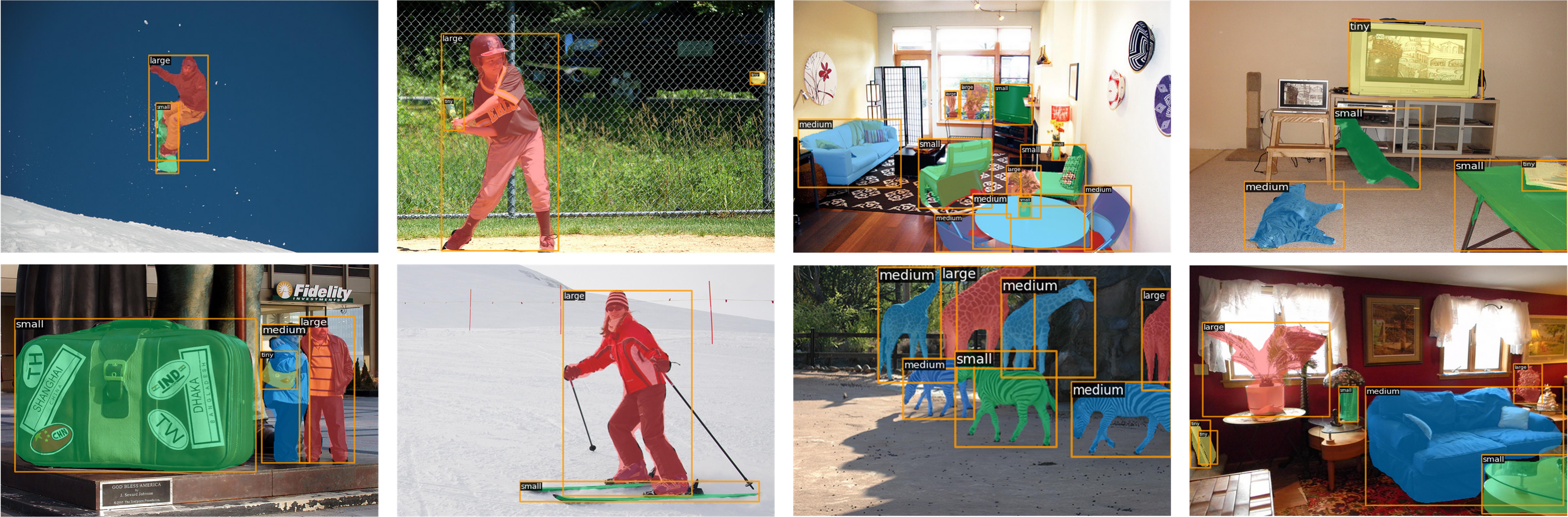}\\
		\vspace{-0.7em}
		\caption{Mask selection results by DynaMask. Each color corresponds to a candidate resolution (\textcolor{red}{red}$\to$large, \textcolor{blue}{blue}$\to$medium, \textcolor{green}{green}$\to$small, \textcolor{yellow}{yellow}$\to$tiny).}
		\label{fig7}
		\vspace{-1em}
	\end{figure*}

\noindent\textbf{Comparison with state-of-the-art methods.}
We then compare the segmentation performance of DynaMask with many other state-of-the-art methods on COCO. The results are listed in Tab.~\ref{tab6}. All models are trained on COCO \texttt{train2017} and evaluated on COCO \texttt{test-dev2017}. Without bells and whistles, DynaMask surpasses these methods with visible margins. Furthermore, we compare our method with other representative two-stage IS methods on Cityscapes \texttt{val} set in Tab.~\ref{tab7}. Our method outperforms Mask R-CNN~\cite{he2017mask} by 4.2\%. Noticeably, its performance on large objects is 5.8\% AP higher than Mask R-CNN. This is because DynaMask employs high-resolution masks to achieve high-quality segmentation on large and difficult objects. RefineMask~\cite{zhang2021refinemask} also produces outstanding performance since it refines the mask prediction with fine-grained features, but it costs many computations due to the multi-stage refinement.

\noindent\textbf{Visualizations of segmentation results.} In Fig.~\ref{fig6}, we visualize the segmentation results of DynaMask together with two representative methods: Mask R-CNN~\cite{he2017mask}, and PointRend~\cite{kirillov2020pointrend}. One can see that our method achieves finer and more accurate predictions around the objects' boundaries. This is because Dynamask introduces an r-FPN to fuse complementary information from multiple stages, so that difficult instances can be segmented with large masks which contain more fine-grained details. Mask R-CNN~\cite{he2017mask} employs a uniform coarse mask for prediction, while PointRend~\cite{kirillov2020pointrend} makes point-wise refinement only at the blurred regions, which is not enough to capture sufficient details. 


\subsection{Ablation Study}
\noindent\textbf{Mask resolution prediction.} To better understand how DynaMask selects suitable masks for different instances, we show the mask selection results in Fig.~\ref{fig7}. As can be seen, the ``hard'' objects with irregular shapes and complex boundaries are assigned large masks, such as the ``potted plant'', ``person", ``giraffe", \etc. On the contrary, the ``easy'' samples with regular shapes and less details are assigned small masks, such as the ``skis", ``suitcase", ``dining table" and so on. It is worth mentioning that the choice of mask size is independent of object size. For example, in the third image, the small but hard object ``potted plant" is allocated a large mask, while the large but easy sample ``suitcase'' in the fifth image can be accurately segmented with a small mask. We provide quantitative analysis of the correlations between the predicted mask resolutions and the class in \textbf{supplemental files}.


\begin{table*}[!t]
	\begin{minipage}[htbp]{\textwidth}
		\begin{minipage}[t]{0.35\textwidth}
			\centering
			\makeatletter\def\@captype{table}\makeatother  
			\scalebox{0.82}{				
				\begin{tabular}{c|c|ccc}
					\hline \rowcolor{gray!20}
					Method      &  AP & $\rm AP_S$ & $\rm AP_M$ & $\rm AP_L$ \\ \hline \hline
					Mask R-CNN~\cite{he2017mask}   &  33.8  &  12.0  &  31.5  &  51.8\\
					
					PointRend~\cite{kirillov2020pointrend}   &  35.8  &  -  & - & - \\
					BMask R-CNN~\cite{cheng2020boundary}    &  35.8  &  -  & - & - \\
					RefineMask~\cite{zhang2021refinemask}   &  37.6  &  14.6  & 34.0 & 58.1 \\
					DynaMask  &  38.0  &    14.8 & 35.1 & 57.6 \\ \hline
				\end{tabular}
			}\\
			\vspace{-0.8em}
			\caption{The results on Cityscapes \texttt{val} set with ResNet-50-FPN backbone.}
			\label{tab7}
		\end{minipage}
		\hspace{1.0em}
		\begin{minipage}[t]{0.25\textwidth}
			\centering
			\makeatletter\def\@captype{table}\makeatother
			\scalebox{0.75}{		
				\setlength{\tabcolsep}{0.5em}
				\begin{tabular}{c|c|c|c|c}
					\hline \rowcolor{gray!20}
					$\mathcal{C}_t$ & $\lambda_2$ & FLOPs & $\rm \Delta$ & AP   \\ \hline \hline
					1.0      & 0.3         & 1.27G & -9\%   & 37.6 \\
					1.0      & 0.4         & 1.13G & -19\%  & 37.6 \\ 
					1.0      & 0.5         & 1.06G & -24\%  & 37.5\\
					\hline
					0.8      & 0.4         & 0.92G & -34\% &  37.4\\
					0.6      & 0.4         & 0.83G & -41\% &  37.2\\ 
					0.4      & 0.4         & 0.64G & -54\% &  36.8\\ \hline
				\end{tabular}
				\hspace{-0.2em}
				\setlength{\tabcolsep}{0.2em}
				
			}
			\vspace{-0.8em}
			\caption{The influence of the budget constraint.}
			\label{tab3}
		\end{minipage}
		\hspace{0.8em}
		\begin{minipage}[t]{0.35\textwidth}
			\centering
			\makeatletter\def\@captype{table}\makeatother  
			\scalebox{0.75}{				
				\begin{tabular}{c|c|c|c}
					\hline \rowcolor{gray!20}
					Method       & AP   &  FLOPs  & FPS  \\ \hline \hline
					Mask R-CNN~\cite{he2017mask}   & 34.7 &  0.5G  & 12.4           \\
					PointRend~\cite{kirillov2020pointrend}    & 35.6 &  0.9G  & 9.4      \\
					DCT-Mask$^\dag$~\cite{shen2021dct}     & 36.5 &  0.5G  & 11.8             \\
					HTC~\cite{chen2019hybrid}          & 37.4 &  -     & 3.9             \\ \hline
					DynaMask & 37.6 &  1.4G      & 8.3          \\ 
					DynaMask ($\mathcal{C}_t=0.4$) & 36.8 &  0.64G  &  11.2            \\ 
					\hline
				\end{tabular}
			}
			\vspace{-0.8em}
			\caption{Speed comparison of different methods on \texttt{val} set, $\dag$: multi-scale training.}
			\label{tab4}
		\end{minipage}
	\end{minipage}

\end{table*}

\begin{table*}[!t]
	\begin{minipage}[htbp]{\textwidth}
		\begin{minipage}[t]{0.5\textwidth}
			\centering
			\makeatletter\def\@captype{table}\makeatother  
			\scalebox{0.75}{				
				\begin{tabular}{c|c|c|cc|ccc}
					\hline \rowcolor{gray!20}
					Mask Size & FLOPs & AP  & $\rm AP_{50}$ & $\rm AP_{75}$ & $\rm AP_{S}$ & $\rm AP_{M}$ & $\rm AP_{L}$ \\ \hline\hline
					14$\times$14     &   0.23G    &  32.9  &  55.3  &  34.4  &  19.3  &  35.8  &  43.6  \\ \hline
					28$\times$28     &   0.62G   &  36.1  &   57.1  &  38.7  &  20.4  &  39.2  &  48.2  \\ \hline
					56$\times$56     &   1.01G    &  37.1  &  57.3  &  40.0  &  20.5  &  40.0  &  49.7  \\ \hline
					112$\times$112   &   1.40G    &  37.6  &  57.4   &  40.5  &  20.7  &  40.4  &  50.3  \\ \hline
				\end{tabular}
			}
			\vspace{-0.8em}
			\caption{Performance of using different mask sizes.}
			\label{tab2}
		\end{minipage}
		\hspace{0.3em}
		\begin{minipage}[t]{0.5\textwidth}
			\centering
			\makeatletter\def\@captype{table}\makeatother
			\scalebox{0.75}{		
				\setlength{\tabcolsep}{0.5em}
				\begin{tabular}{c|c|c|cccc}
					\hline \rowcolor{gray!20}
					Mask Size    & FLOPs & AP & Tiny & Small & Medium & Large \\ \hline \hline
					28$\times$28 & 0.62G &  36.1 & 0	&1	&0	&0   \\
					\small{Size-based} & 0.56G &  35.9 & 47\%	&32\%	&14\%	&8\%       \\
					\small{DynaMask ($\mathcal{C}_t=0.4$)}& 0.64G & 36.8 & 35\%  &  34\%	 &  21\%	& 10\%      \\
					DynaMask     & 1.4G  & 37.6 & 0	&0	&0	&1        \\ \hline
				\end{tabular}
				\hspace{-0.2em}
				\setlength{\tabcolsep}{0.2em}
				
			}
			\vspace{-0.8em}
			\caption{Comparison with size-based mask selection method.}
			\label{tab9}
		\end{minipage}
	\end{minipage}
	\vspace{-1.0em}
\end{table*}

\noindent \textbf{Influence of the budget constraint.}  In Tab.~\ref{tab3}, we explore the influence of budget constraint on the model complexity (FLOPs). The average FLOPs are calculated by taking the average over all instances of the validation set. By tuning hyper-parameters $\mathcal{C}_t$ and $\lambda_2$ to different values, we get models of different computational costs. For example, the FLOPs are reduced by about 19\% without sacrificing the segmentation performance by setting $\mathcal{C}_t$ and $\lambda_2$ to 1.0 and 0.4, respectively. This demonstrates that there exists much redundancy when using the large mask ($112\times 112$) for all instances. Actually, many instances in COCO dataset can be efficiently segmented with smaller masks. Thus the redundant computation for easier samples can be reduced, while the accuracy of harder samples can be maintained by still using larger masks. By setting the target FLOPs $\mathcal{C}_t$ to a smaller value (\textit{e.g.}, 0.4, 0.6, 0.8), more significant computation reduction can be achieved at a slight degradation of accuracy. For instance, by setting both $\mathcal{C}_t$ and $\lambda_2$ to 0.4, more than half of the FLOPs (around 54\%) are reduced while still producing competitive segmentation results (36.8\% AP). 

\noindent\textbf{Speed comparison of different methods.} To validate the efficiency of our model, we compare the model accuracy, FLOPs, and runtime of different two-stage IS methods in Tab.~\ref{tab4}. Inference time is tested on one NVIDIA TITAN RTX with the input
size $800\times 1333$. Compared to these methods, our DynaMask method achieves visible performance gain at a small amount of extra computational cost. Specifically, DynaMask ($\mathcal{C}_t=0.4$) outperforms PointRend~\cite{kirillov2020pointrend} and DCT-mask~\cite{shen2021dct} by 1.2\% AP and 0.3\% AP, respectively, with comparable runtime. Though HTC~\cite{chen2019hybrid} produces a similar segmentation result (0.2\% lower AP) to DynaMask, it is two times slower than DynaMask because it performs hybrid cascade refinement on both detection and segmentation tasks, resulting in a large amount of memory and computation overhead.

\noindent\textbf{Influence of mask size.} The Mask Switch Module (MSM) outputs four different candidate mask resolutions $[14\times14,28\times28,56\times56,112\times112]$. We choose one of them as the uniform output mask size and report the corresponding results in Tab.~\ref{tab2}. As can be seen, models with larger mask resolutions could achieve higher segmentation performance, especially on large objects, but the computational cost also increases obviously. For example, the performance is improved by 3.2\%, 4.2\%, and 4.7\% AP when the mask resolution is increased from $14\times 14$ to $28\times 28$, $56\times 56$ and $112\times 112$, respectively. Nevertheless, the performance tends to saturate as the mask size increases further. 

\noindent\textbf{Size-based mask selection method.}
We compare the performance and mask distributions of different mask selection methods in Tab.~\ref{tab9}. The baseline is performed by using a uniform mask size ($28\times 28$) for all objects. Size-based method denotes assigning masks according to the size of object. Specifically, we assign a mask $\hat{m}_i^k$ to the instance of width $w$ and height $h$ by the following rule:
{\small \begin{align}
	k=\left \lfloor k_0+log_2(\sqrt{wh}/\sqrt{w_0h_0} ) \right \rfloor, 
	\label{eq8}
\end{align}}
where $w_0$ and $h_0$ denote the width and height of the input image, respectively. and $k_0$ denotes the index of the highest mask resolution $112\times 112$, \textit{i.e.}, $k_0=4$. Intuitively, Eq.~\ref{eq8} means that the larger object will be assigned a higher-resolution mask. This conforms to our common sense, since the larger objects usually contain many details which need finer-grained masks to achieve high-quality prediction. 

As can be seen from Tab.~\ref{tab9}, the size-based method attains comparable performance with the baseline at a lower cost. Our DynaMask ($\mathcal{C}_t=0.4$) outperforms the baseline by 0.7\% AP and the size-based method by 0.9\% AP at a slight extra computation cost, demonstrating that the proposed mask selection strategy could better partition different objects according to their segmentation difficulties and assign more suitable masks to achieve better performance. More analyses about the distribution of predicted mask scales can be found in \textbf{supplemental materials}.  

%
\vspace{-0.5em}
\section{Conclusion}
\vspace{-0.5em}
In this work, we proposed a simple yet effective method to improve instance segmentation performance at a small amount of computation and memory overhead. We devised a dual-level FPN structure for better exploring complementary contextual and detailed information from multiple pyramid levels. Specifically, in addition to traditional image-level FPN (i-FPN), we augmented a region-level top-down path (r-FPN) to gradually enlarge the mask size and incorporate more details from i-FPN. Furthermore, to reduce the computation and memory cost led by using large masks, we introduced a Mask Switch Module (MSM) to adaptively select suitable masks for each proposal so that the redundant computation for easy samples can be reduced by using smaller masks. Extensive experimental results demonstrated that our method achieved significant performance gains with moderate extra computation overhead. 

\clearpage
\section{Supplemental Materials}
In this section, we provide the following materials:
\begin{itemize}
	\item[$\bullet$] Analyses on mask resolution prediction (\cf Sec4.3-Mask Resolution Prediction in the main paper);
	\item[$\bullet$] Distribution of predicted mask resolutions (\cf Sec4.3-Size-based Mask Selection in the main paper); 
	\item[$\bullet$] Results on LVIS dataset;
	\item[$\bullet$] Comparisons with other variants of FPN;
	\item[$\bullet$] Qualitative results (\cf Sec4.2-Visualizations of Segmentation Results in the main paper).
\end{itemize}

\subsection{Analyses on Mask Resolution Prediction}
\begin{figure}[!h]
	\centering
	\vspace{1em}
	\includegraphics[scale=0.3]{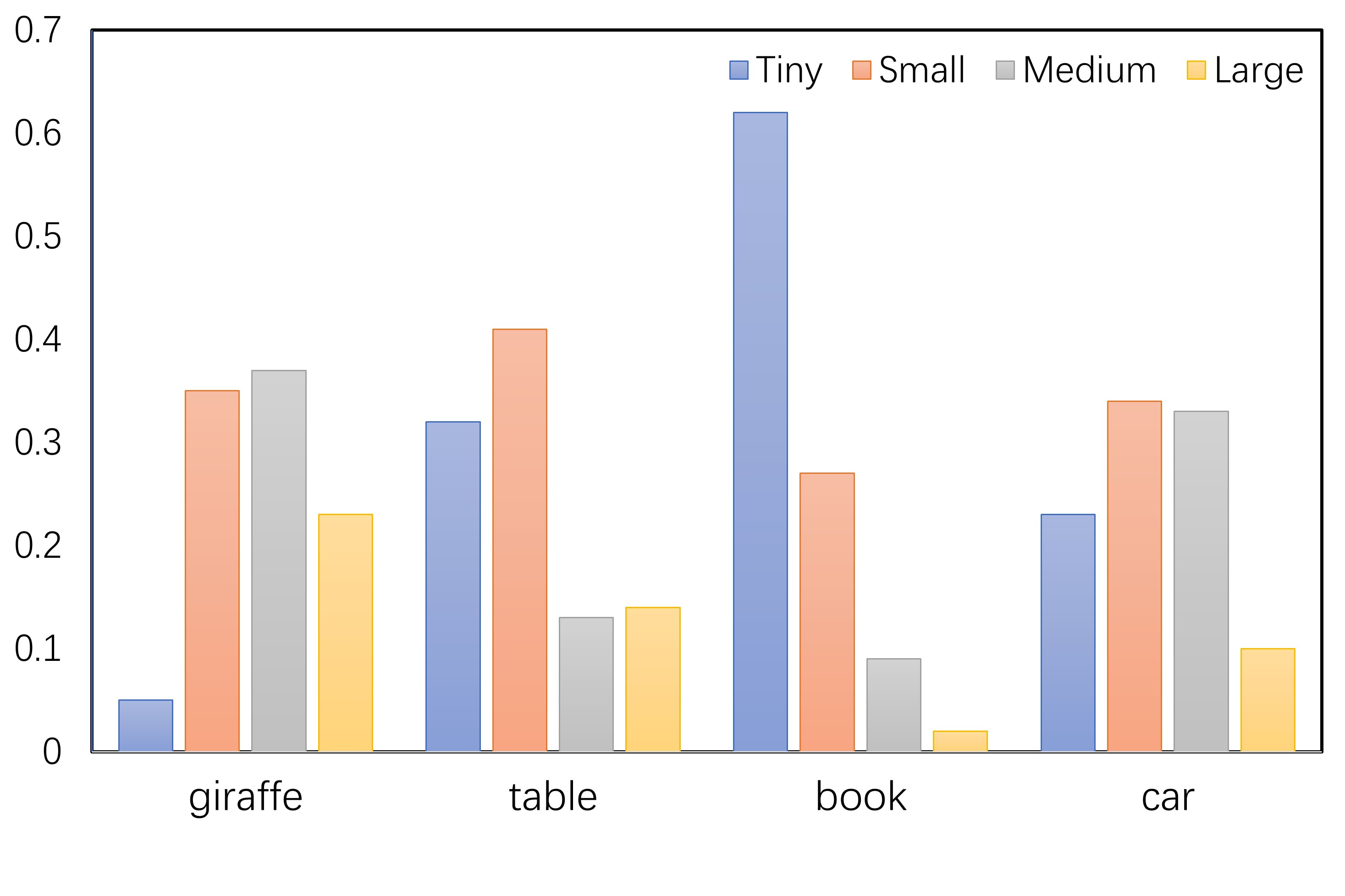}\\
	\vspace{-1.5em}
	\caption{The mask distributions of different classes.}
	\label{correlation}
\end{figure}

{\bf Correlation between mask resolution and class.} We show the mask distributions of different classes for DynaMask ($54\%\downarrow$) in Fig.~\ref{correlation}. As can be seen, the classes with irregular shapes (\eg, ``giraffe") tend to be assigned larger masks, while the classes of regular shapes (\eg, ``book") are  allocated smaller masks for efficiency.

\begin{figure*}[!h]
	\centering
	\includegraphics[scale=0.63]{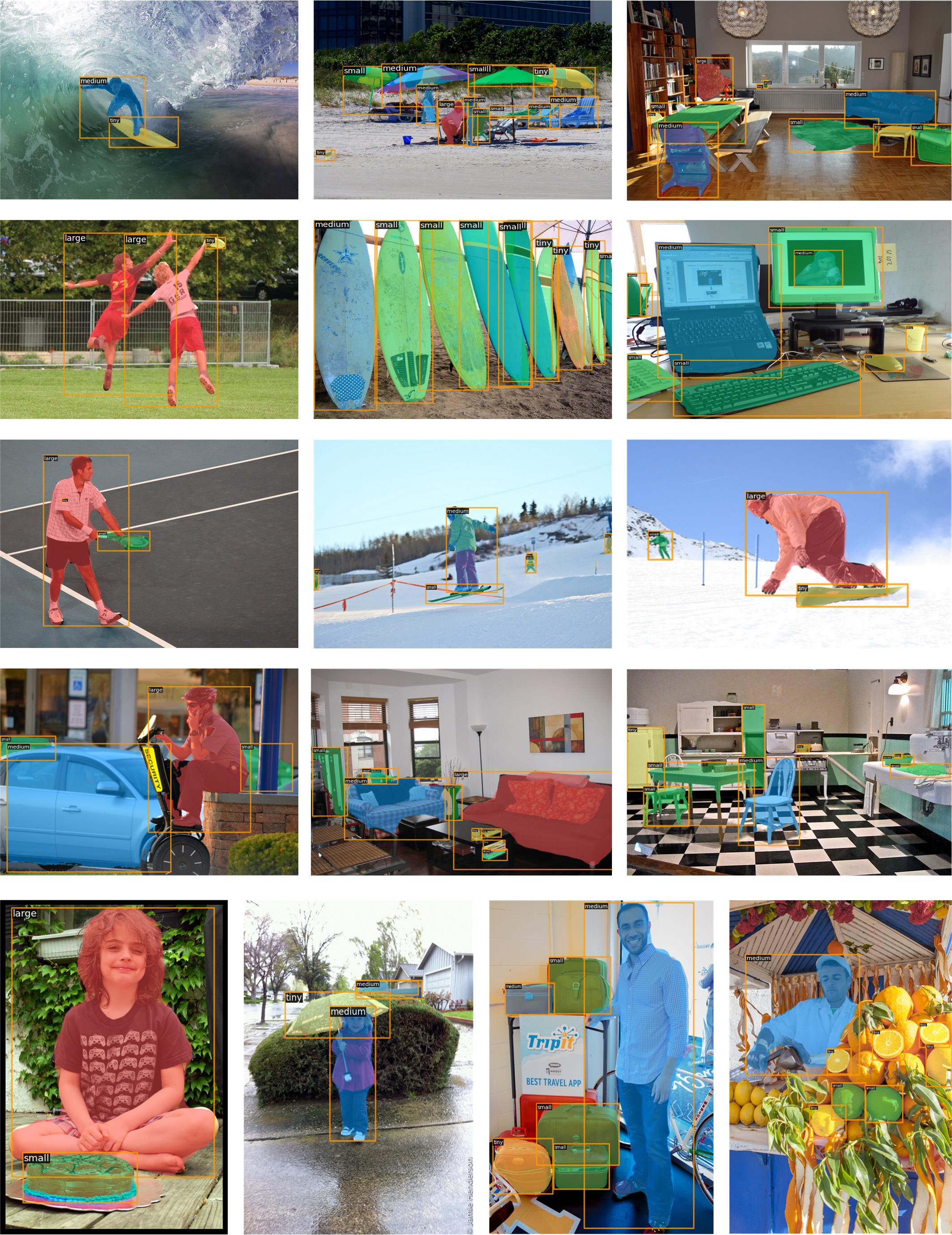}\\
	\caption{Mask selection results by DynaMask. Each color corresponds to a candidate resolution (\textcolor{red}{red}$\to$large, \textcolor{blue}{blue}$\to$medium, \textcolor{green}{green}$\to$small, \textcolor{yellow}{yellow}$\to$tiny).}
	\label{selection}
\end{figure*}

{\bf Mask selection results.} Fig.~\ref{selection} shows more examples of mask resolution selection. As can be seen, the ``hard'' objects with irregular shapes and complex boundaries are assigned larger masks, such as the ``potted plant'', ``person", \etc. On the contrary, the ``easy'' samples with regular shapes and less details are assigned smaller masks, such as the ``skis", ``suitcase", ``dining table" and so on.  

\subsection{Distribution of Predicted Mask Resolutions}
To better understand why our method can reduce the computations, we count the number of masks assigned to different resolutions and report them in Tab.~\ref{distribution}. We take ``Size-Based" approach, which assigns mask according to the size of objects (please refer to Eq.~8 in the main paper), as a reference here. One can see that by selecting lower-resolution masks, the FLOPs can be reduced by 19\% without performance degradation, demonstrating that it is redundant to use large masks for all instances. DynaMask ($54\%\downarrow$) outperforms Size-Based approach by 0.9\% AP with comparable FLOPs, indicating that the proposed dynamic mask selection is more effective to select suitable masks.

\begin{table}[!h]
	\centering
	\scalebox{0.8}{
		\begin{tabular}{c|p{0.5cm}p{0.5cm}p{0.5cm}p{0.6cm}|c|c}
			\hline \rowcolor{gray!20}
			Method          & {\small Tiny} & {\small Small} & {\small Med} & {\small Large} & FLOPs & AP   \\ \hline \hline
			Size-Based & 0.47 & 0.32  & 0.14   & 0.08  & 0.56G & 35.9 \\
			DynaMask  & 0 & 0  & 0   & 1  & 1.4G & 37.6 \\
			DynaMask ($19\% \downarrow$) & 0.05 & 0.13  & 0.25   & 0.56  & 1.13G & 37.6 \\
			DynaMask ($54\% \downarrow$) & 0.35 & 0.34  & 0.21   & 0.10  & 0.64G & 36.8 \\ \hline
	\end{tabular}}
	\caption{The distribution of predicted masks.}
	\label{distribution}
\end{table}

\subsection{Results on LVIS Dataset}
Since the quality of mask annotations in COCO~\cite{lin2014microsoft} is limited, which makes it a less convincing candidate for validating the accuracy of high-resolution mask prediction. We hence evaluate the performance on LVIS~\cite{gupta2019lvis}, which has higher-quality annotations. The corresponding results are reported in Tab.~\ref{lvis} below. DynaMask outperforms Mask R-CNN~\cite{he2017mask} by 3.6\% AP since we apply higher-resolution masks to ``harder" samples, which usually contain more details.   
\begin{table}[!h]
	\centering
	\scalebox{0.8}{
		\begin{tabular}{c|c|c|ccc}
			\hline \rowcolor{gray!20}
			Method     & Backbone & AP   & AP$_r$ & AP$_c$ & AP$_f$ \\ \hline \hline
			Mask R-CNN~\cite{he2017mask} & R50-FPN  & 22.1 & 10.1  & 21.7  & 31.7  \\
			DynaMask   & R50-FPN  & 25.7 & 13.9  & 24.4  & 32.0      \\ \hline
	\end{tabular}}
	\caption{Results on LVIS validation dataset.}
	\label{lvis}
\end{table}

\subsection{Comparisons with Other Variants of FPN}
We adopt the Dual-FPN to integrate region-level features for refining mask predictions. We compare our method with other variants of FPN, such as PANet~\cite{liu2018path}, BiFPN~\cite{ghiasi2019fpn}, and NAS-FPN~\cite{tan2020efficientdet}. The results are reported in Tab.~\ref{fpn}. One can see that our method performs better since we adaptively aggregate complementary information from different feature levels.
\begin{table}[!h]
	\centering
	\scalebox{0.8}{
		\begin{tabular}{c|c|cc|ccc}
			\toprule\rowcolor{gray!20}
			method & AP & AP$_{50}$ & AP$_{75}$ & AP$_{S}$ & AP$_{M}$ & AP$_{L}$ \\ \hline\hline
			PANet~\cite{liu2018path}   &  35.9  &  56.8  & 38.3 & 19.4 & 39.2 & 49.0 \\
			BiFPN~\cite{ghiasi2019fpn}  & 36.4   &  57.2  & 39.0 & 19.2 & 39.8 & 49.5  \\
			NAS-FPN~\cite{tan2020efficientdet} & 36.0  & 56.9  & 38.7 & 18.9 & 38.5 & 49.9 \\
			DynaMask (ours) &  37.6  & 57.4 & 40.5 & 20.7 & 40.4 & 50.3 \\
			\bottomrule
	\end{tabular}}
	\caption{Comparisons with different variations of FPN.}
	\label{fpn}
\end{table}

\subsection{Qualitative Results}
\begin{figure*}[!h]
	\centering
	\includegraphics[scale=0.33]{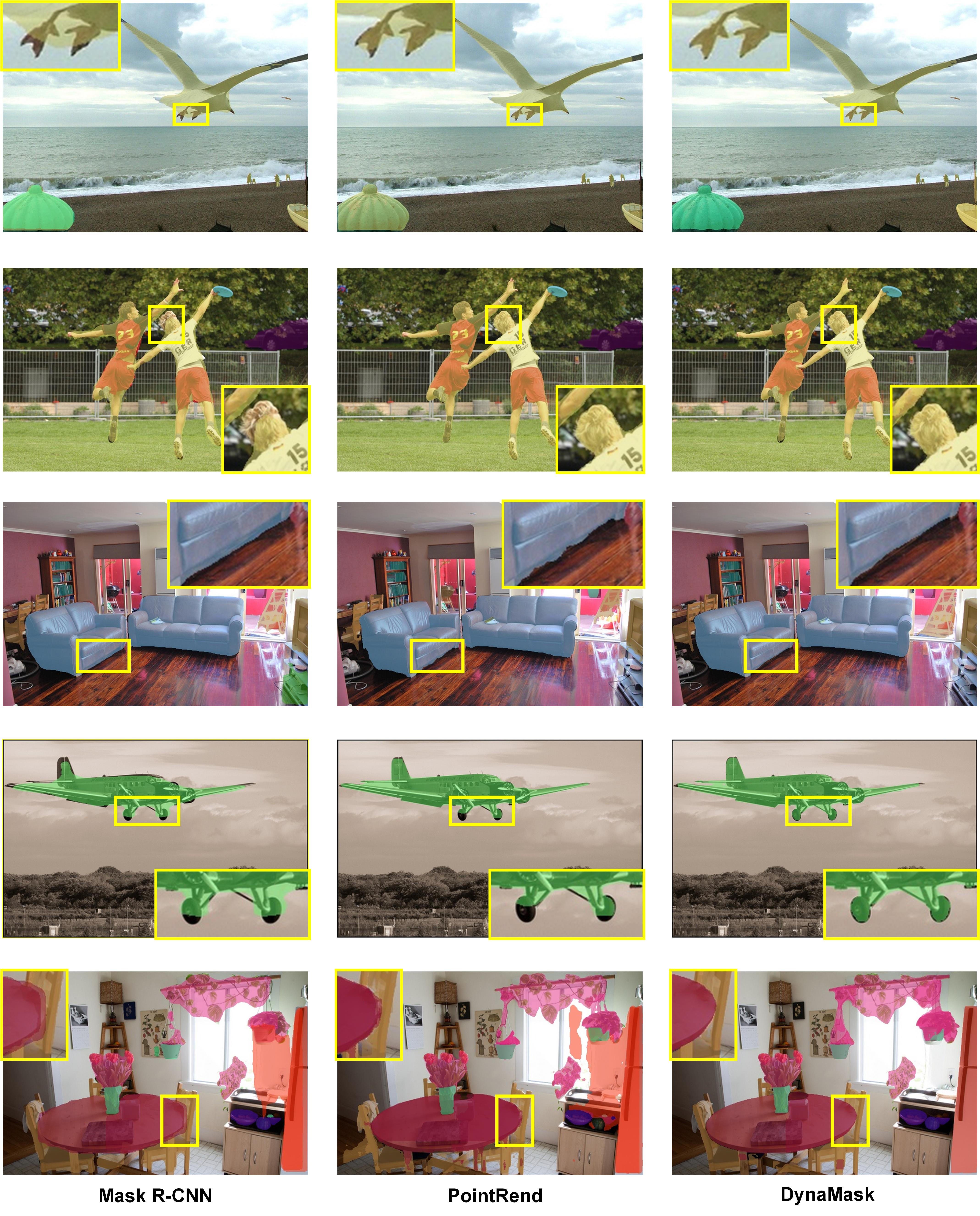}\\
	\caption{Qualitative comparisons between Mask R-CNN~\cite{he2017mask}, PointRend~\cite{kirillov2020pointrend}, and DynaMask, using ResNet-50-FPN backbone.}
	\label{vis}
\end{figure*}

\begin{figure*}[!h]
	\centering
	\includegraphics[scale=0.33]{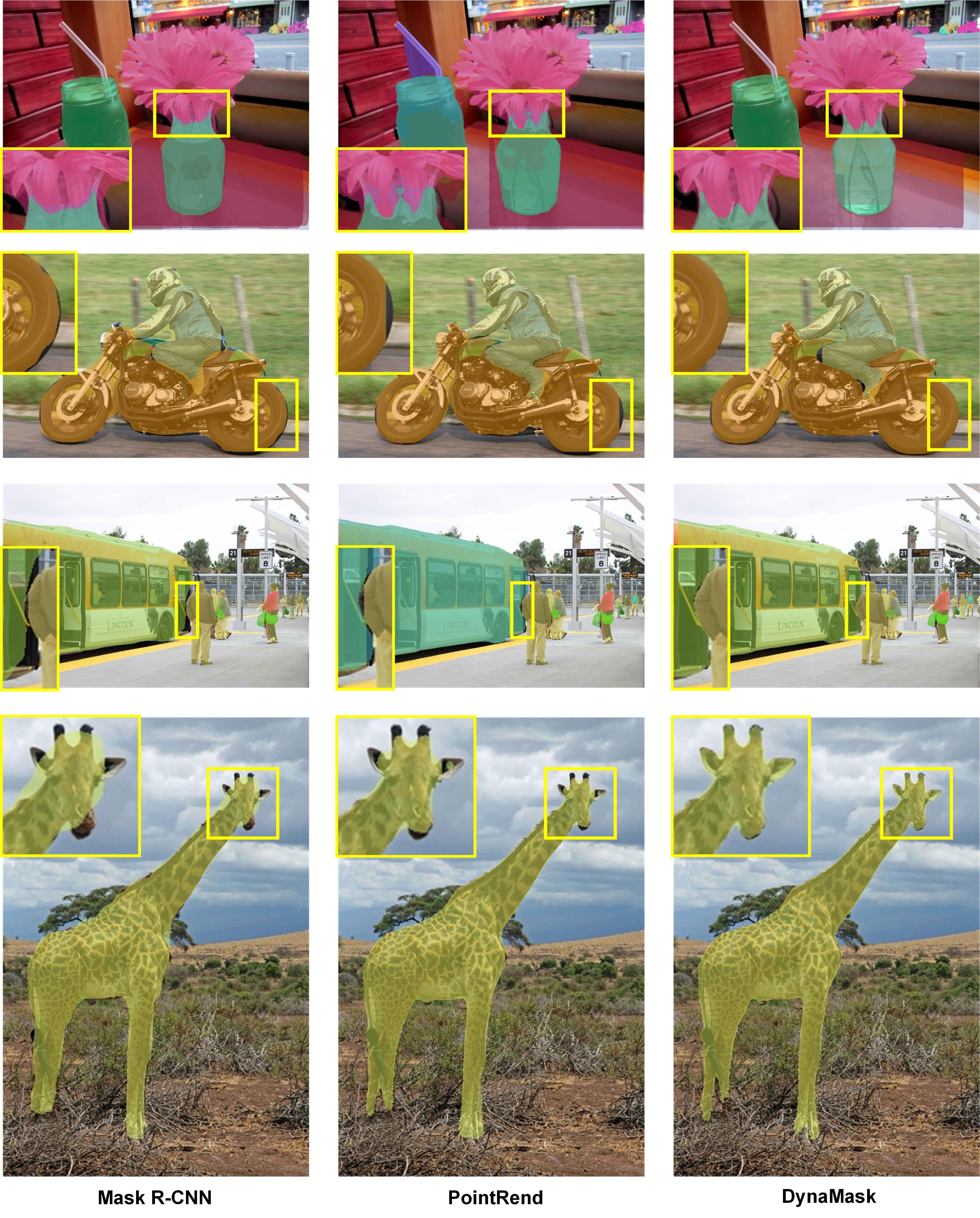}\\
	\caption{Qualitative comparisons between Mask R-CNN~\cite{he2017mask}, PointRend~\cite{kirillov2020pointrend}, and DynaMask, using ResNet-50-FPN backbone.}
	\label{vis2}
\end{figure*}

\begin{figure*}[!h]
	\centering
	\includegraphics[scale=0.33]{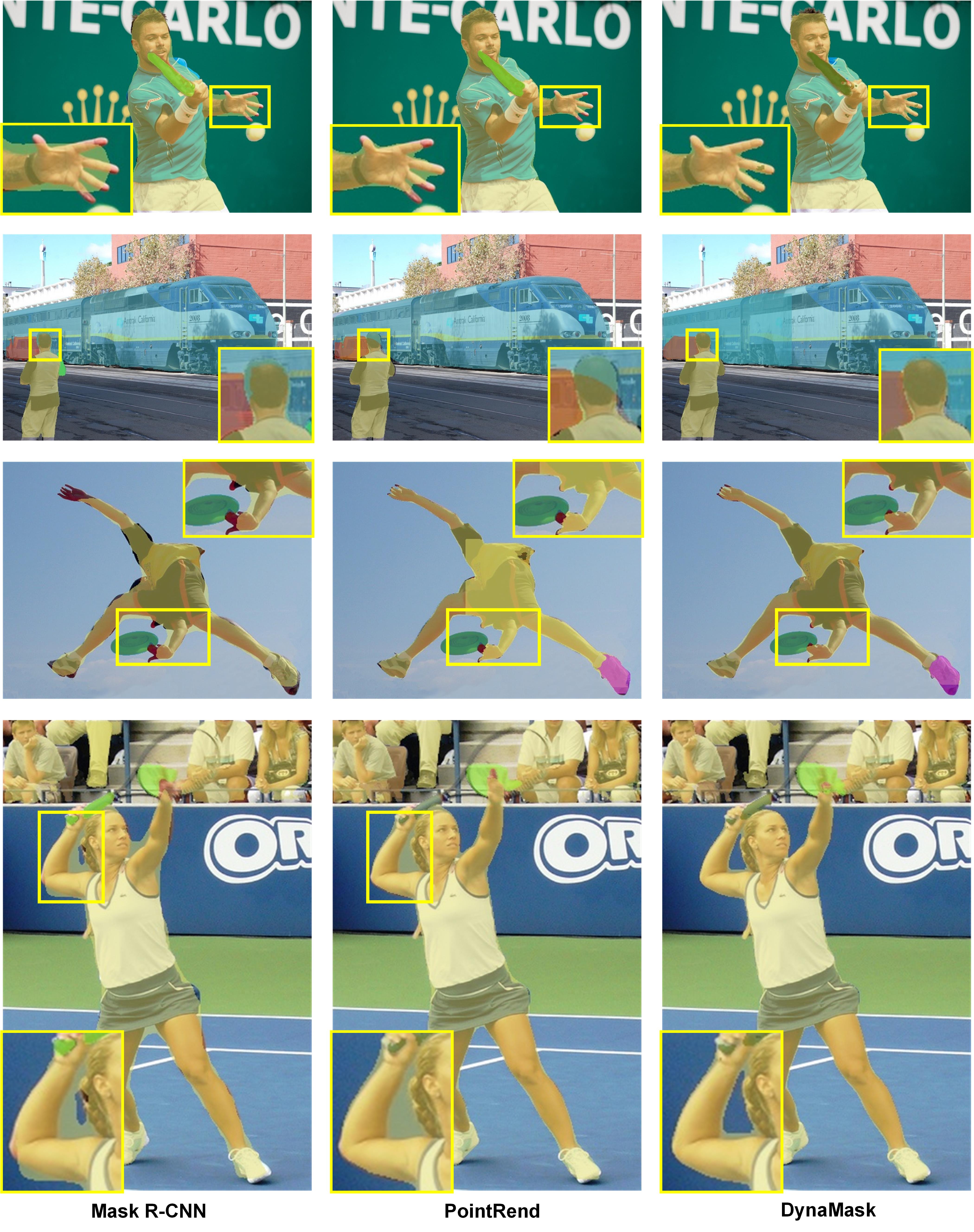}\\
	\caption{Qualitative comparisons between Mask R-CNN~\cite{he2017mask}, PointRend~\cite{kirillov2020pointrend}, and DynaMask, using ResNet-50-FPN backbone.}
	\label{vis3}
\end{figure*}

In Figs.~\ref{vis}-\ref{vis3}, we visualize the segmentation results of DynaMask together with two representative methods: Mask R-CNN~\cite{he2017mask} and PointRend~\cite{he2017mask}. As can be seen, our method achieves finer and more accurate predictions, especially on the boundary regions of objects. 

{\small
\bibliographystyle{ieee_fullname}
\bibliography{egbib}
}

\end{document}